\documentclass[10pt,twocolumn,letterpaper]{article}

\usepackage[pagenumbers]{cvpr} 

\usepackage{graphicx}
\usepackage{amsmath}
\usepackage{amssymb}
\usepackage{booktabs}

\usepackage{fontawesome}
\usepackage[linesnumbered,ruled,vlined]{algorithm2e}
\usepackage{tabularx}
\usepackage{multirow}
\usepackage[table,xcdraw]{xcolor}
\usepackage{subcaption}

\usepackage[pagebackref,breaklinks,colorlinks]{hyperref}

\usepackage[capitalize]{cleveref}
\crefname{section}{Sec.}{Secs.}
\Crefname{section}{Section}{Sections}
\Crefname{table}{Table}{Tables}
\crefname{table}{Tab.}{Tabs.}

\def\cvprPaperID{*****} 
\def\confName{CVPR}
\def\confYear{2022}

\begin{document}

\title{SIn-NeRF2NeRF: Editing 3D Scenes with Instructions through Segmentation and Inpainting \\ Performance Improvement Track}

\author{Jiseung Hong\\
KAIST\\
Computer Science\\
{\tt\small jiseung.hong@kaist.ac.kr}
\and
Changmin Lee\\
KAIST\\
Computer Science\\
{\tt\small lcm914@kaist.ac.kr}
\and
Gyusang Yu\\
KAIST\\
Mechanical Engineering\\
{\tt\small peter.yu@kaist.ac.kr}
}
\maketitle


\begin{figure}[h]
    \centering
    \includegraphics[width=\columnwidth]{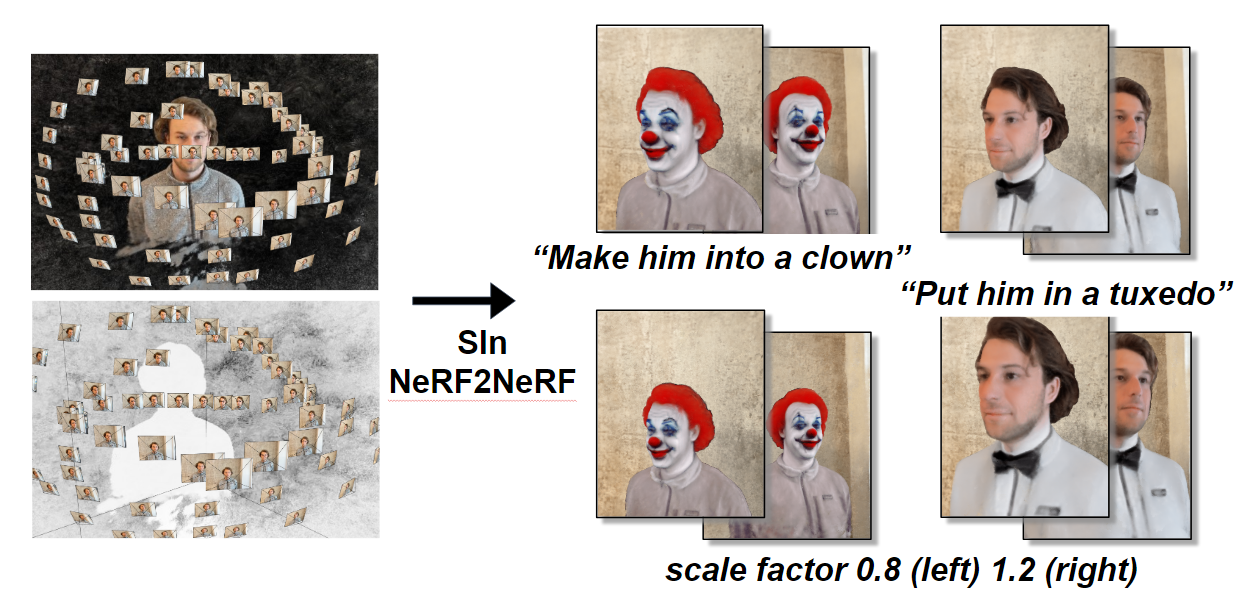}
    \caption{\textbf{Overview}: We propose the method SIn-NeRF2NeRF, enabling wide range of object edition including translation, rotation, and scale changes.}
    \label{fig:teaser}
\end{figure}

\begin{abstract}
\textbf{TL;DR} Perform 3D object editing selectively by disentangling it from the background scene.

Instruct-NeRF2NeRF~\cite{in2n} (in2n) is a promising method that enables editing of 3D scenes composed of Neural Radiance Field (NeRF)~\cite{nerf} using text prompts. However, it is challenging to perform geometrical modifications such as shrinking, scaling, or moving on both the background and object simultaneously. In this project, we enable geometrical changes of objects within the 3D scene by selectively editing the object after separating it from the scene. We perform object segmentation and background inpainting respectively, and demonstrate various examples of freely resizing or moving disentangled objects within the three-dimensional space. Code is available at: \href{https://github.com/KAISTChangmin/SIn-NeRF2NeRF}{https://github.com/KAISTChangmin/SIn-NeRF2NeRF}
\end{abstract}

\section{Introduction}
\label{sec:intro}

The current leading method for representing 3D scenes is Neural Radiance Fields (NeRF), which can generate realistic novel views from a sparse set of images, given the camera parameters. The ability to freely and stably edit such 3D scenes is one of the most critical technologies for implementing the real world in VR/AR applications.

The introduced Instruct-NeRF2NeRF is a powerful tool that allows humans to edit the scene itself based on given input. Additionally, SPIn-NeRF\cite{spinnerf} has created a background scene by interactively removing selected objects from the entire scene. InpaintNeRF360~\cite{wang2023inpaintnerf360} is a recent work that appropriately fuses the concepts of both. Unlike SPIn-NeRF, which is limited to inpainting results of frontal scenes, InpaintNeRF360 demonstrated the ability to perform inpainting across 360-degree scenes using human prompts.

The 3D scene edition results from Instruct-NeRF2NeRF shows promising results in geometrical addition and minor perturbations, while some of the updates in objects were not reflected in the background scene. For instance, when an object goes through geometrical shrinking, the original part of the background wouldn’t fill the perturbed part of the object.

In this project, our objective is to achieve a refined object edition by separating the object from the scene, restoring the original portion of the background without the object through inpainting, and selectively editing the object. Ultimately, this approach enables the disentanglement of the object and background, facilitating more precise and effective modifications in 3D scene reconstruction.

Therefore, we propose SIn-NeRF2NeRF (sn2n) that modifies Instruct-NeRF2NeRF (in2n) based on the code of SPIn-NeRF to perform object edition. The key challenge lied in the modification of in2n framework to get RGBA image as input, and yield the object scene with no background. We elaborate the detailed implementation of this and other processes in the subsequent sections. We verify the fidelity of our model based on the datasets provided by in2n and SPIn-NeRF. Furthermore, we confirm its functionality on a custom dataset, thereby demonstrating its robustness and scalability.


\begin{figure*}
    \centering
    \includegraphics[width=\textwidth]{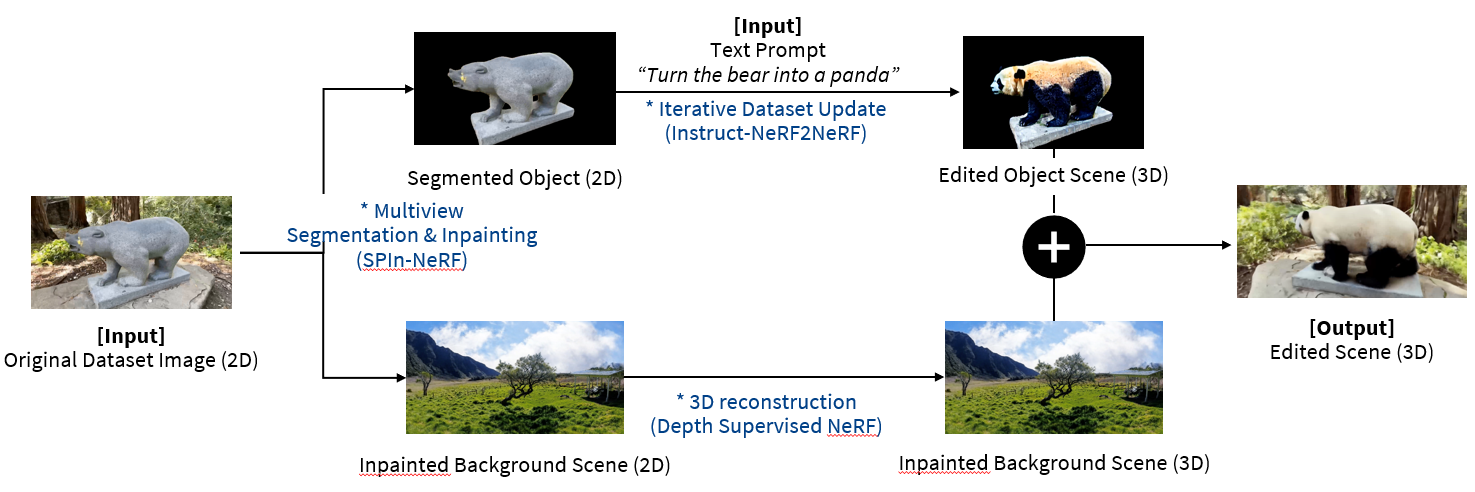}
    \caption{\textbf{Main Framework}: Our method processes an input consisted of original NeRF scene and a text prompt, yielding an edited NeRF scene with the disentangled object. Detailed information regarding the implementation can be found in Section~\ref{sec:implementation}.}
    \label{fig:overview}
\end{figure*}\label{sec:method}
\section{Method Summary}

As shown in Figure~\ref{fig:overview}, sn2n receives images of a scene from multiple views and object masks as input. Additionally, it takes a text prompt as an input to output an edited 3D scene where the object is modified. During this process, techniques such as multiview segmentation and SPIn-NeRF were utilized to disentangle the object and the background.

\subsection{Problem Setup}
    As specified above, sn2n takes a text prompt and a NeRF scene as inputs to separate the scene into object and background components. For the object, it performs scene editing and scaling, while for the background, it applies 3D inpainting, and then merges the two scenes. In the process of editing the object scene, it employs a method $I_{object, i+1}^v \leftarrow U_{\theta}(I_{object, i}^v, t; I_{object, 0}^v, c^{T})$, for an unedited object image $I_{object, 0}^v$, a text instruction $c^{T}$, and a noisy input $z_{t}$, where $z_{0} = \varepsilon(I_{object, i}^{v})$. Here, $U_{\theta}$ is defined as DDIM sampling, consistent with the definition in in2n. Meanwhile, for the background scene, a 3D inpainting method called SPIn-NeRF was used. For further details in SPIn-NeRF, refer to Section~\ref{sec:spinnerf}. Consequently, we merge two NeRF scenes where the object is disentangled from the background.
    
\subsection{Baseline Inspection}
\label{sec:baseline inspection}
\begin{figure}[ht]
        \centering
        \begin{subfigure}[b]{0.11\textwidth}
            \includegraphics[width=\textwidth]{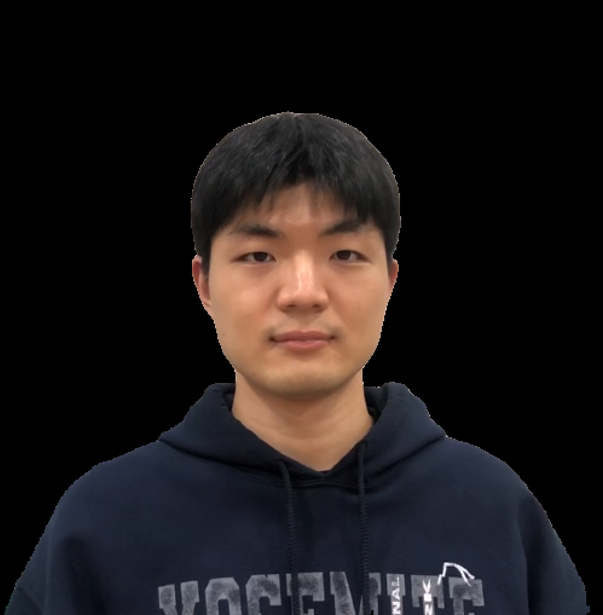}
            \caption{Original}
            \label{fig:image1}
        \end{subfigure}
        \hfill
        \begin{subfigure}[b]{0.11\textwidth}
            \includegraphics[width=\textwidth]{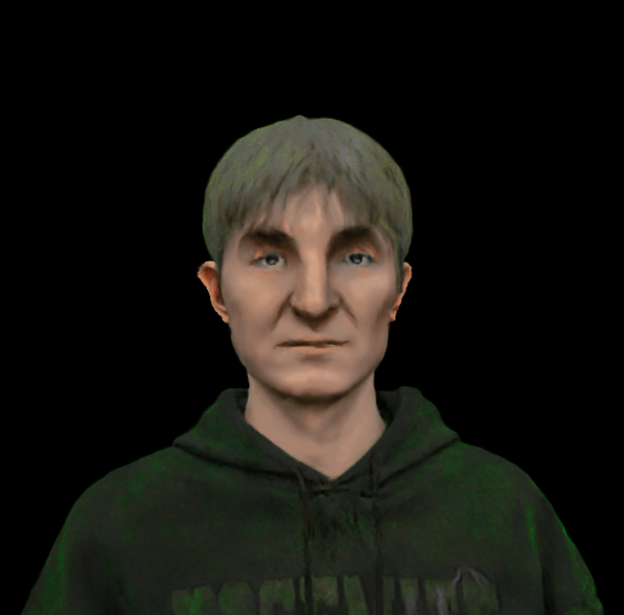}
            \caption{Iter-5000}
            \label{fig:image2}
        \end{subfigure}
        \hfill
        \begin{subfigure}[b]{0.11\textwidth}
            \includegraphics[width=\textwidth]{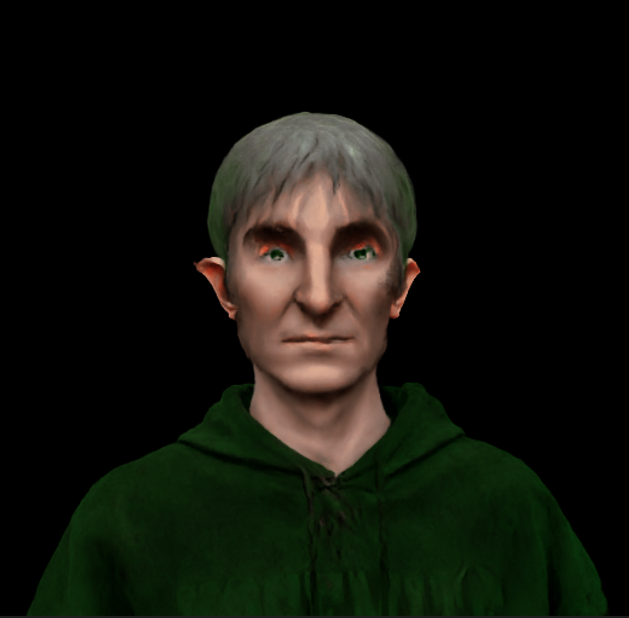}
            \caption{Iter-10000}
            \label{fig:image3}
        \end{subfigure}
        \hfill
        \begin{subfigure}[b]{0.11\textwidth}
            \includegraphics[width=\textwidth]{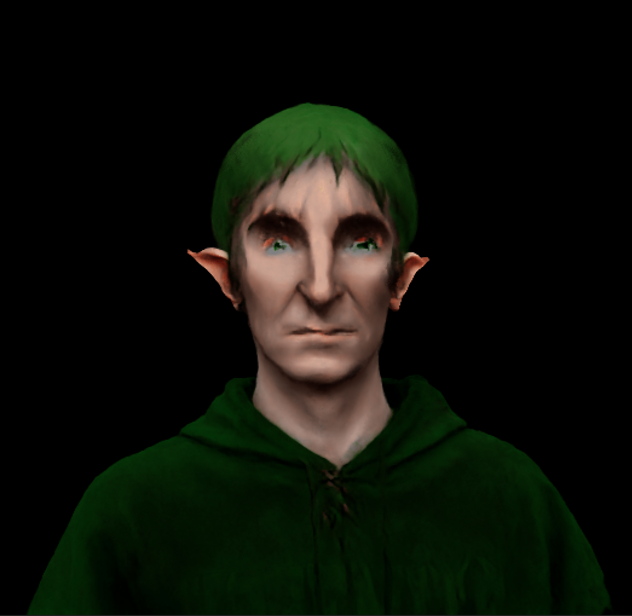}
            \caption{Iter-15000}
            \label{fig:image4}
        \end{subfigure}
        \caption{Iterative Dataset Update (IDU)}
        \label{fig:idu}
    \end{figure}

    Instruct-NeRF2NeRF is a novel work performing object editing within NeRF scenes and serves as the baseline method for our project. In the Section~\ref{sec:result}, we compare the outcomes of applying in2n after separately learning the object and background scene using the SPIn-NeRF framework, against when in2n applied to the original scene.
    Instruct-NeRF2NeRF takes a typical NeRF scene as input and follows an iterative dataset update pipeline to update the scene. Per each iteration step, this dataset update pipeline takes the novel scene image with added random gaussian noise and prompt from the rendered view as input, and outputs an image updated via InstructPix2Pix\cite{ip2p} (ip2p).
    Initially the difference between the text prompt and the original scene is huge, and 2D images reconstructed according to the prompt turns out 3D inconsistent. However as learning continues, the scene gradually changes starting from aspects across views. This can be seen in Figure~\ref{fig:idu}. Iterative dataset update is showing incremental changes to (such as pointy ears, blue eyes, green t-shirts, etc.) per iteration. 

    \vspace{-1em}

\subsection{SPIn-NeRF}
\label{sec:spinnerf}
    SPIn-NeRF plays a pivotal role in our project by inpainting the background of the original scene with sparse object selection. SPIn-NeRF starts with training the given NeRF scene and acquires disparity maps for novel views. Consequently, the the object mask is given as input and it perform LaMA inpainting in every training views. Additional LaMa inpainting is done to the initially acquired disparity map. Finally the background scene is constructed using these two inpainted rgb image and disparity inpainted image sets.

\section{Implementation Details}
\label{sec:implementation}
    We have implemented the entire process from data preparation/preprocessing (2D image multiview segmentation) to applying in2n on the object scene and synthesizing the 3D object scene with the 3D inpainted background scene. We borrowed 3D inpainting of the background scene algorithm from LaMa~\cite{LaMa}.  The detailed methods and algorithms applied at each step are as follows.

\subsection{2D Image Multiview Segmentation}
    We implemented code that allows interactive 2D image multiview segmentation using the Segment Anything Model (SAM)~\cite{SAM}. It takes the original 2D image set as input, separates the object and background by providing sparse annotations for each frame, and retrieves the object mask.

\subsection{Object Scene Reconstruction \& Applying in2n}
\label{sec:in2n}
\begin{figure}[ht]
        \centering
        \begin{minipage}[b]{0.22\linewidth}
            \includegraphics[width=\linewidth]{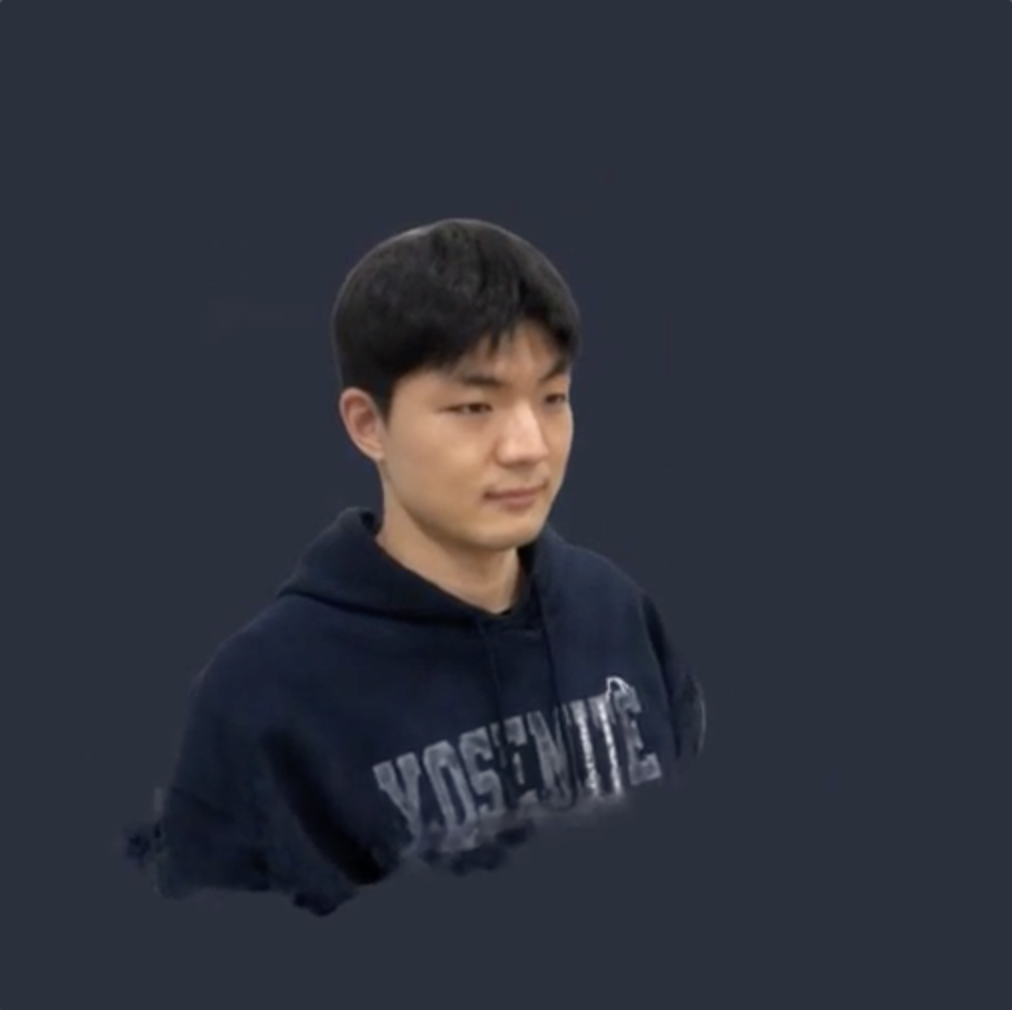}
        \end{minipage}
        \hspace{0.5mm}
        \begin{minipage}[b]{0.22\linewidth}
            \includegraphics[width=\linewidth]{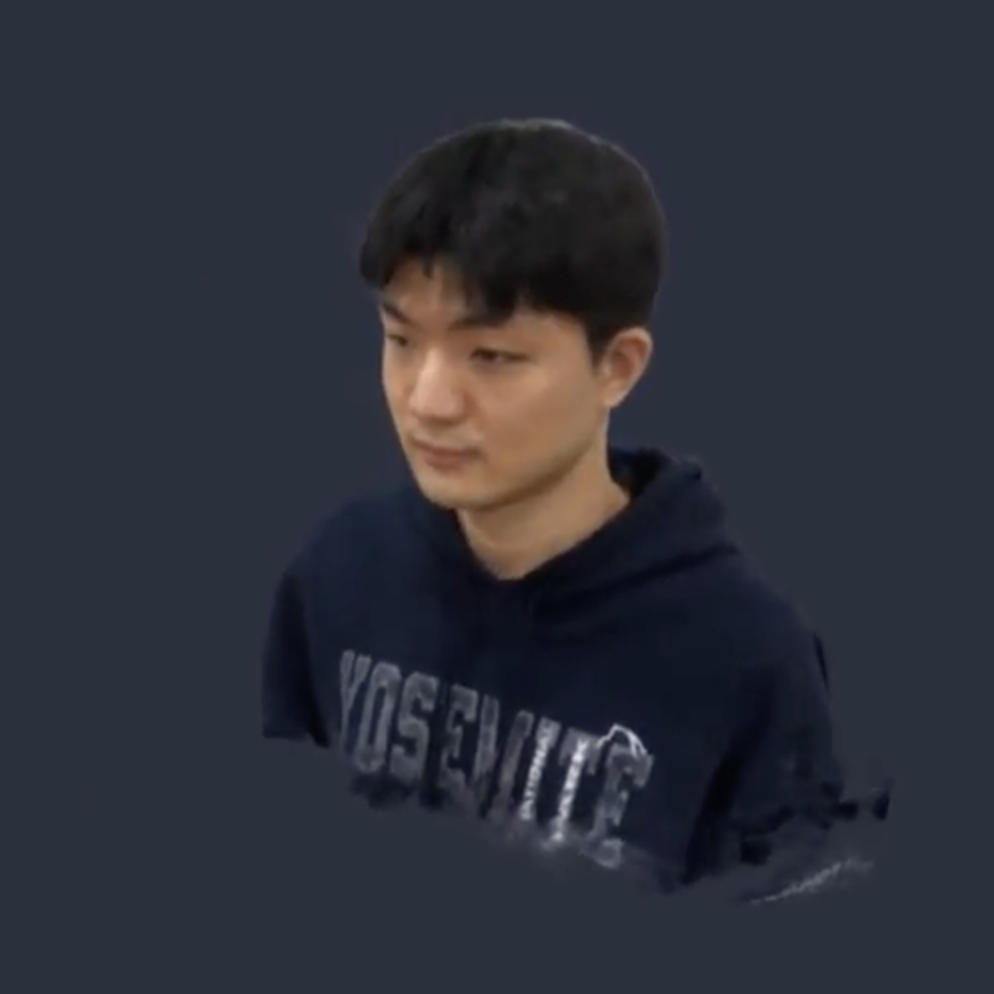}
        \end{minipage}
        \hfill
        \begin{minipage}[b]{0.22\linewidth}
            \includegraphics[width=\linewidth]{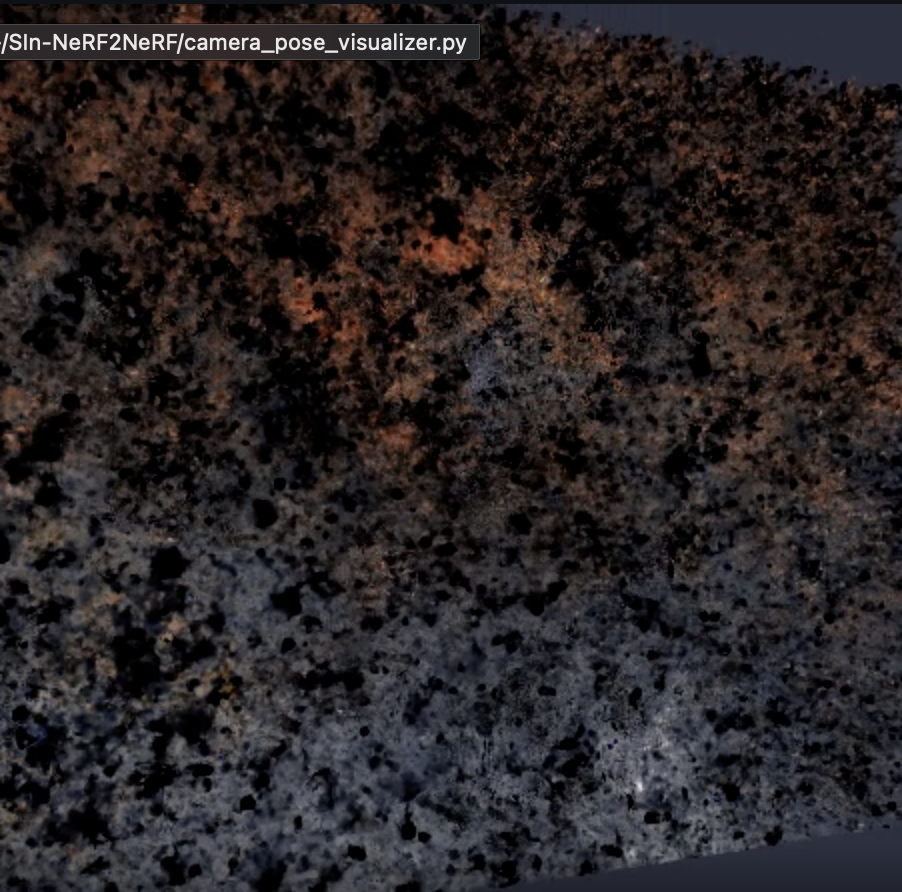}
        \end{minipage}
        \hspace{0.5mm}
        \begin{minipage}[b]{0.22\linewidth}
            \includegraphics[width=\linewidth]{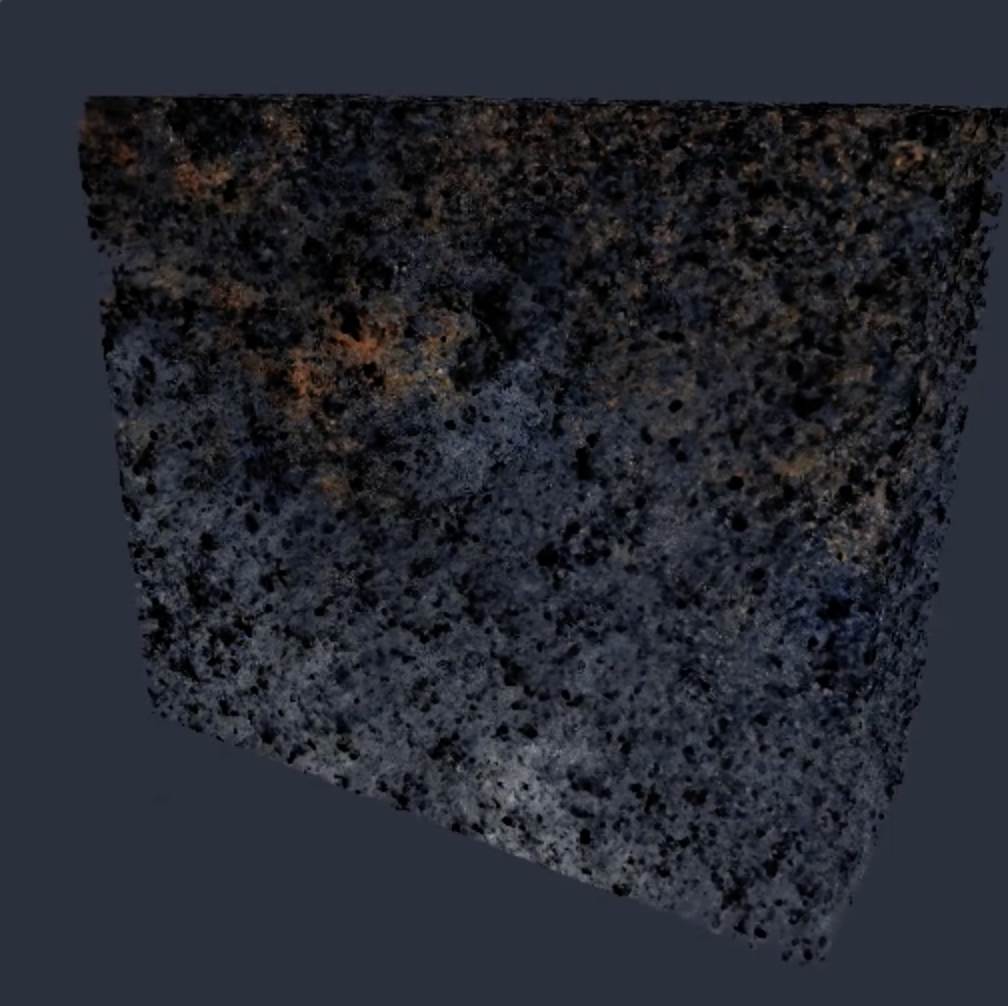}
        \end{minipage}
    \caption{Object scene with (left two) and without (right two) the random background color method.}
    \label{fig:rbc}
    \end{figure}

    For the object scene, we train a DSNeRF~\cite{DSNeRF} using an RGBA image set concatenated from the object mask and object image. During this process, we implement the random background color technique borrowed from the instant-ngp~\cite{instant-ngp} code to ensure effective training of the object scene with a transparent background. The random background color technique involves alpha blending each view with a random color. Figure~\ref{fig:rbc} shows significant difference in results compared to when technique is not applied.
    

    We develop a code to edit the object scene based on text prompts by porting the in2n pipeline, especially the Iterative Dataset Update (IDU) algorithm, from Nerfacto to DSNeRF. The IDU algorithm used in our team's code is specified in Algorithm~\ref{alg:in2n_alg}. Our novelty lies in the manipulation of object scenes, since we modify the in2n framework to get input of RGBA images, as opposed to the conventional RGB images. As mentioned, alpha blending is applied and outputs the object scene within the RGBA images, with no background in 3D space when reconstructed. The modified framework preserves the integrity of the original format while stably incorporate our targeted modifications.

    \begin{algorithm}[h]
    \SetAlgoLined
    Dataset $\leftarrow$ original images \( I_{0}^{v} \) for viewpoints \( v \)\;
    \For{each iteration}{
        Randomly order viewpoints \( v \)\;
        \For{\( d \) image updates}{
            \For{each viewpoint in ordered \( v \)}{
                1. Alpha blend RGBA image with black background (RGBA $\rightarrow$ RGB)\;
                2. Update image using ip2p\;
                3. Segment the object (RGB $\rightarrow$ RGBA)\;
            }
        }
        \For{\( n \) NeRF updates}{
            Sample random rays from entire training dataset\;
            Update NeRF with a mix of old and newly updated images\;
        }
    }
    \caption{Iterative Dataset Update Process}
    \label{alg:in2n_alg}
    \end{algorithm}

\subsection{Background Scene Reconstruction}
    Using the code provided by SPIn-NeRF, we train the DSNeRF for the inpainted background scene. We perform 2D inpainting on the original 2D background images, where the object part is segmented out, using LaMa, and then trained the 3D inpainted background scene using RGB loss, disparity loss, LPIPS loss, and true depth loss.

\subsection{3D NeRF Scene Synthesis}
    The method for synthesizing the edited object scene from Section~\ref{sec:in2n} with the inpainted background scene using SPIn-NeRF was adopted from ml-neuman~\cite{ml-neuman}. Since both scenes share the same camera parameters, we sort the sampled points for the same rays generated in the object and background scenes by their depth values $Z$ as in Equation~\ref{eq:z}.

    \begin{equation}
        P_{\text{sorted}} = \left\{ t_i \;|\; t_i \in P_{\text{object}} \cup P_{\text{bkg}} \text{ and } Z(t_i) \text{ is sorted} \right\}
    \label{eq:z}
    \end{equation}
    

\subsection{Object Transformation}
    The process of transforming (scaling, translation, and rotation) a disentangled object scene from the background commences with the utilization of COLMAP to ascertain the coordinates $P$ of the 3D object. Subsequently, these coordinates are transformed around the centroid $O$ using a transformation factor $scale$, $rotate$, and $trans$, thereby acquiring the 3D coordinates of the transformed object $P^{'}$. This transformation doesn't change the actual 3D model but alters how it is perceived from the camera's viewpoint during the rendering process. 


\section{Experimental Results}
\label{sec:result}
    We confirm the robustness of our framework by dataset provided by SPIn-NeRF team and in2n team. We focus primarily on presenting qualitative results, complementing these with quantitative results using CLIP metric. The face scene is mainly presented due to its intuitiveness. The method can be also applied for other datasets. We train our Segmented NeRF (object scene) for 2k iterations, which takes about 30 minutes on a single RTX 4060 Ti. Running multiview inpainter (background scene) also takes about 30 minutes on a same device.

\subsection{Qualitative Results}
\begin{figure}[ht]
        \centering
        \begin{subfigure}[b]{0.11\textwidth}
            \includegraphics[width=\textwidth]{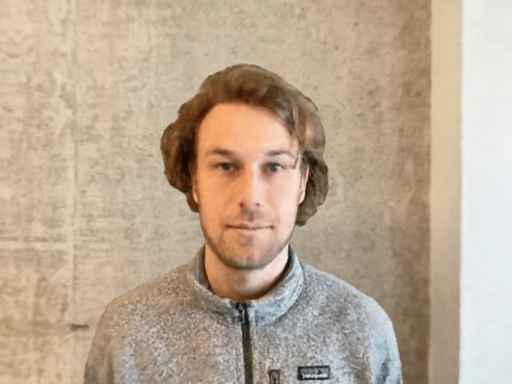}
            \caption{Original}
            \label{fig:image1}
        \end{subfigure}
        \hspace{1mm}
        \begin{subfigure}[b]{0.11\textwidth}
            \includegraphics[width=\textwidth]{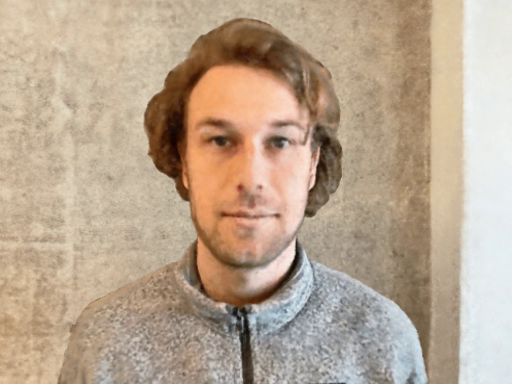}
            \caption{Scaling}
            \label{fig:image2}
        \end{subfigure}
        \hspace{0mm}
        \begin{subfigure}[b]{0.11\textwidth}
            \includegraphics[width=\textwidth]{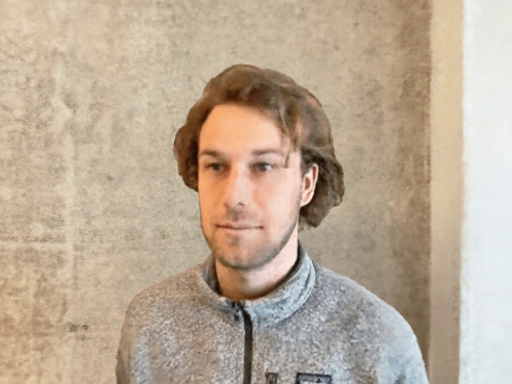}
            \caption{Rotation}
            \label{fig:image3}
        \end{subfigure}
        \hspace{0mm}
        \begin{subfigure}[b]{0.11\textwidth}
            \includegraphics[width=\textwidth]{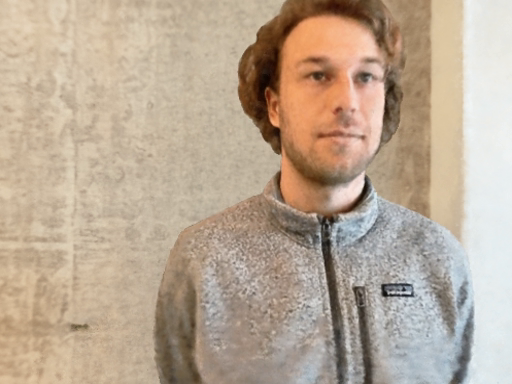}
            \caption{Translation}
            \label{fig:image4}
        \end{subfigure}
        \caption{\textbf{Qualitative Results}: Object Transformation.}
        \label{fig:qual}
    \end{figure}

    The advantage of our method is that by separately training the object and background and then merging them, it allows for not only editing through prompt input but also free adjustment of position and scaling. In Figure~\ref{fig:qual}, we can observe the free panning of an object in the scene.

    \vspace{-1.3em}
    
\subsubsection*{Baseline Comparison}
\begin{figure}[ht]
        \centering
        \begin{minipage}[b]{0.11\textwidth}
            \centering
            \raisebox{1.0\height}{
                \begin{tabular}{@{}c@{}}
                    \textit{"Make him into}\\
                    \textit{a clown"}
                \end{tabular}
            }
        \end{minipage}
        \hfill
        \begin{subfigure}[b]{0.105\textwidth}
            \includegraphics[width=\textwidth]{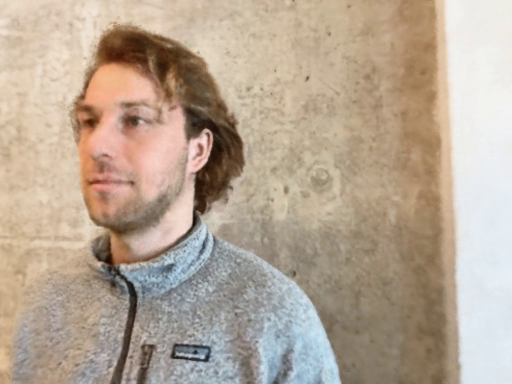}
        \end{subfigure}
        \begin{subfigure}[b]{0.105\textwidth}
            \includegraphics[width=\textwidth]{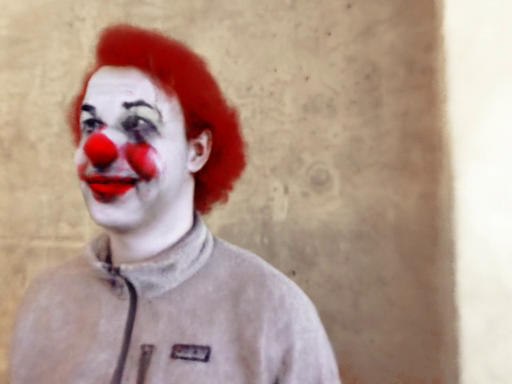}
        \end{subfigure}
        \begin{subfigure}[b]{0.105\textwidth}
            \includegraphics[width=\textwidth]{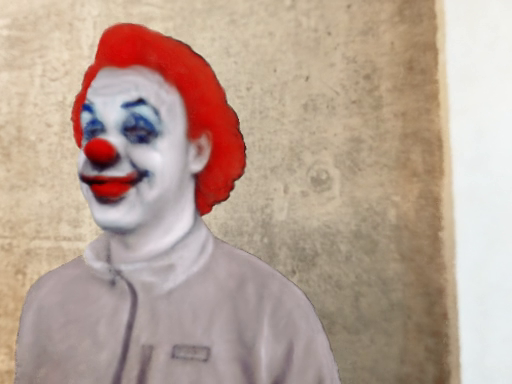}
        \end{subfigure}

        \begin{minipage}[b]{0.11\textwidth}
            \centering
            \raisebox{1.0\height}{
                \begin{tabular}{@{}c@{}}
                    \textit{"Put him in a}\\
                    \textit{Tuxedo"}
                \end{tabular}
            }
            \label{fig:text}
        \end{minipage}
        \hfill
        \begin{subfigure}[b]{0.105\textwidth}
            \includegraphics[width=\textwidth]{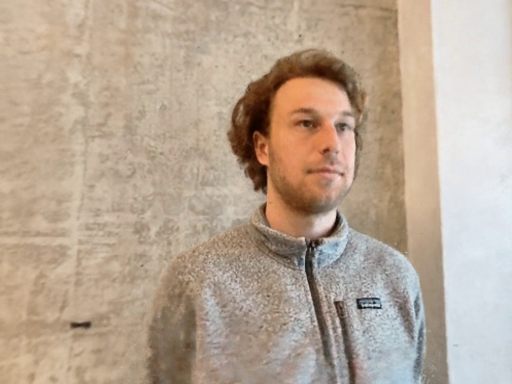}
            \caption{Original}
            \label{fig:image2}
        \end{subfigure}
        \begin{subfigure}[b]{0.105\textwidth}
            \includegraphics[width=\textwidth]{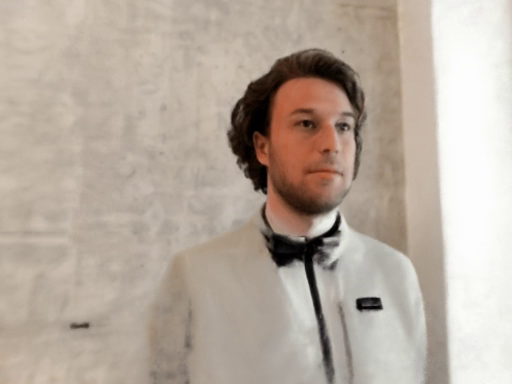}
            \caption{in2n}
            \label{fig:image3}
        \end{subfigure}
        \begin{subfigure}[b]{0.105\textwidth}
            \includegraphics[width=\textwidth]{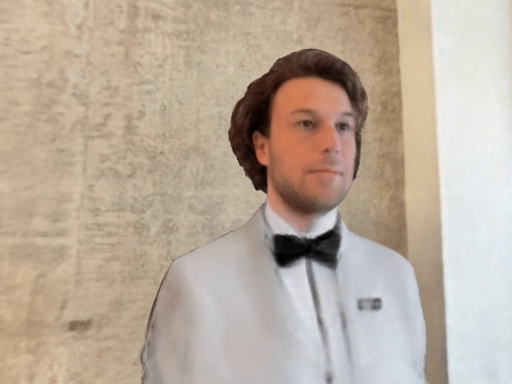}
            \caption{sn2n}
            \label{fig:image4}
        \end{subfigure}
        \caption{Comparison between in2n \& sn2n}
        \label{fig:base1}
    \end{figure}

    It can be confirmed from Figure~\ref{fig:base1} that in2n has been stably implemented within the SPIn-NeRF framework. When comparing the results of sn2n with in2n for various prompts, it can be observed that the outcomes are similar because both employ the same instruct pix2pix's 2D diffusion image editing. However, in the detailed aspects below, it is evident that sn2n has a geometric advantage over in2n, such as smoother background reconstruction.

    \vspace{-1em}
    
\subsection{Quantitative Results}
    As mentioned in the baseline paper\cite{in2n}, editing is a subjective task, so the focus should be on qualitative results rather than quantitative results when measuring performance improvement. However, as in ip2p and in2n, we analyze quantitatively in two ways. The results of comparing sn2n to the baseline model in2n based on the alignment of 3D edits with text instructions and the temporal consistency of edits across views are shown in Table~\ref{tab:clip}. We evaluated the metrics based on the face scene with the text prompt \textit{"Make him into a clown"} and \textit{"Put him in a Tuxedo"}.

    \begin{table}[!hbt]
    \begin{tabular}{|
    >{\columncolor[HTML]{FFFFFF}}l |
    >{\columncolor[HTML]{FFFFFF}}c 
    >{\columncolor[HTML]{FFFFFF}}c |
    >{\columncolor[HTML]{FFFFFF}}c 
    >{\columncolor[HTML]{FFFFFF}}c |}
    \hline
    \cellcolor[HTML]{FFFFFF}                                                                                             & \multicolumn{2}{c|}{\cellcolor[HTML]{FFFFFF}\begin{tabular}[c]{@{}c@{}}CLIP Text-Image\\ Direction Similarity ↑\end{tabular}} & \multicolumn{2}{c|}{\cellcolor[HTML]{FFFFFF}\begin{tabular}[c]{@{}c@{}}CLIP Direction\\ Consistency ↑\end{tabular}} \\ \cline{2-5} 
    \multirow{-2}{*}{\cellcolor[HTML]{FFFFFF}\begin{tabular}[c]{@{}l@{}}\\ Face Scene \textbackslash\end{tabular}} & \multicolumn{1}{c|}{\cellcolor[HTML]{FFFFFF}in2n}                                    & Ours                                  & \multicolumn{1}{c|}{\cellcolor[HTML]{FFFFFF}in2n}                          & Ours                                 \\ \hline
    $\rightarrow$ Clown                                                                                                          & \multicolumn{1}{c|}{\cellcolor[HTML]{FFFFFF}\textbf{0.2372}}                         & 0.2081                                 & \multicolumn{1}{c|}{\cellcolor[HTML]{FFFFFF}0.9071}                        & \textbf{0.9117}                       \\ \hline
    $\rightarrow$ Tuxedo                                                                                                         & \multicolumn{1}{c|}{\cellcolor[HTML]{FFFFFF}0.0251}                                  & \textbf{0.0481}                        & \multicolumn{1}{c|}{\cellcolor[HTML]{FFFFFF}0.8451}                        & \textbf{0.8599}                       \\ \hline
    \end{tabular}
    \caption{\textbf{Quantitative Results}: We offer quantitative measures to assess how well the edits align with the text and the consistency of subsequent frames in CLIP space.}
    \label{tab:clip}
    \end{table}
    
    The CLIP values of sn2n are comparable to, or even higher than those of in2n. It confirms the robust performance of the scene reconstruction process of our method.

\section{Conclusion}
\label{sec:conclusion}
In this project, we aim to achieve high-quality object editing within 3D scene editing, utilizing interactively segmented input and human prompts. We perform parallel NeRF learning for the received object and scene, where the object is edited using the Instruct-NeRF2NeRF technique as desired, and the scene undergoes inpainting using the SPIn-NeRF method. We demonstrate enhanced object edition (such as translation, scaling, etc.) on people, objects, and large-scale scenes as observed in in2n.

However, the results of object editing significantly varied depending on the performance of iterative dataset update using ip2p. Also, due to the gradual nature of updates, changes are limited to texture and feature modifications rather than dynamic changes like altering the pose. Advancing towards more robust and dynamic object edition capabilities is a direction for future research in this field, and we expect this can be done with RGBA image updating methods like LayerDiffusion~\cite{layerdiff}.

\section*{Acknowledgments}
\label{sec:acknowledgments}
Changmin Lee has implemented the main pipeline of SIn-NeRF2NeRF. Jiseung Hong and Gyusang Yu revised the code and performed qualitative/quantitative analysis. Data acquisition, methodology discussion and report conduction were equally done.

In this project, we borrowed code from SPIn-NeRF, SAM, Colmap, instant-ngp, ML-Neuman, Instruct-NeRF2NeRF, etc., and the details of the borrowed parts are as shown in the Table~\ref{tab:ack}.

\begin{table}[h]
\begin{tabular}{|l|c|}
\hline
SPIn-NeRF          & Multiview inpainting                                \\
\hline
Instruct-NeRF2NeRF & Iterative dataset update                            \\
\hline
SAM                & Interactive object segmentation\\
                   & and mask generation\\
\hline
COLMAP             & Get true depth using\\
& camera parameters              \\
\hline
ML-Neuman          & Merge two different NeRF scenes                     \\
\hline
Instant-ngp        & Random background color\\
\hline       
\end{tabular}
\caption{\textbf{Acknowledgement}: Borrowed codes.}
\label{tab:ack}
\end{table}

{\small
\bibliographystyle{ieee_fullname}
\bibliography{egbib}
}
\end{document}